\documentclass[letterpaper, 10 pt, conference]{ieeeconf} 
\IEEEoverridecommandlockouts
\usepackage{times}
\usepackage{multicol}
\usepackage{multirow}
\usepackage{graphicx}
\usepackage{hhline}
\usepackage{algorithm}
\usepackage{algpseudocode}
\usepackage{amssymb}
\usepackage{amsmath}
\usepackage{dsfont}
\usepackage{booktabs}
\usepackage{tabularx}
\usepackage{soul}
\usepackage{stfloats}
\usepackage{float}
\usepackage{placeins}
\usepackage{graphicx}
\usepackage{amsfonts}
\usepackage{xspace}
\usepackage[T1]{fontenc}
\let\labelindent\relax
\usepackage{enumitem}
\usepackage{comment}
\usepackage{listings}
\usepackage{float}
\usepackage{footnote}
\usepackage[table]{xcolor}
\usepackage{soul}
\usepackage{lipsum}
\usepackage{subcaption}
\usepackage[english]{babel}
\newtheorem{theorem}{Theorem}[section]
\newtheorem{lemma}[theorem]{Lemma}

\usepackage{bbm}

\usepackage[bookmarks=true]{hyperref}
\hypersetup{
    colorlinks=true,
    linkcolor=orange,
    filecolor=orange,      
    urlcolor=orange,
    citecolor=orange,
}
\usepackage[capitalise, nameinlink]{cleveref}

\makeatletter
    \let\NAT@parse\undefined
\makeatother
\usepackage[square, numbers, sort&compress]{natbib}

\crefname{teaser}{Figure}{Figures}
\newcommand{\MyPara}[1]{\noindent\textbf{#1}}

\newcommand{\MethodName}[0]{CAST\xspace}
\newcommand{\ModelName}[0]{CounterfactualVLA\xspace}
\newcommand{\NumEnvs}[0]{3\xspace}
\newcommand{\NumPrompts}[0]{27\xspace}

\newcommand{\DiffCAST}[0]{27}

\newcommand{\ba}{\mathbf{a}}
\newcommand{\bo}{\mathbf{o}}
\newcommand{\Dl}{\mathcal{D}_l}
\newcommand{\Du}{\mathcal{D}_u}
\newcommand{\Da}{\mathcal{D}_a}

\newcommand{\la}{\ell^a}
\newcommand{\lacf}{\ell^{a,cf}}
\newcommand{\lr}{\bar{\ell}}
\newcommand{\lcf}{\ell^{cf}}
\newcommand{\bacf}{\ba^{cf}}
\newcommand{\ta}{t^a}
\newcommand{\pia}{\pi_a}

\definecolor{burntorange}{HTML}{E69138}
\definecolor{berryred}{HTML}{A61C00}
\definecolor{forestgreen}{HTML}{38761D}
\definecolor{skyblue}{HTML}{3C78d8}

\title{\bf \MethodName: Counterfactual Labels Improve Instruction
Following in Vision-Language-Action Models}

\author{
  Catherine Glossop$^{1}$\thanks{Correspondence to \texttt{catherine\_glossop@berkeley.edu}},
  ~William Chen$^{1}$,
  ~Arjun Bhorkar$^{1}$,
  ~Dhruv Shah$^{2}$,
  ~Sergey Levine$^{1}$ \\
$^{1}$ University of California Berkeley ~~~~$^{2}$ Princeton University\\
\texttt{\href{https://cast-vla.github.io}{https://cast-vla.github.io}}\\
}
\begin{document}

\makeatletter
\let\@oldmaketitle\@maketitle
\renewcommand{\@maketitle}{
  \@oldmaketitle
  \vspace{0em}
  \begin{center}
    \vspace{-1em}
    \includegraphics[width=0.85\linewidth]{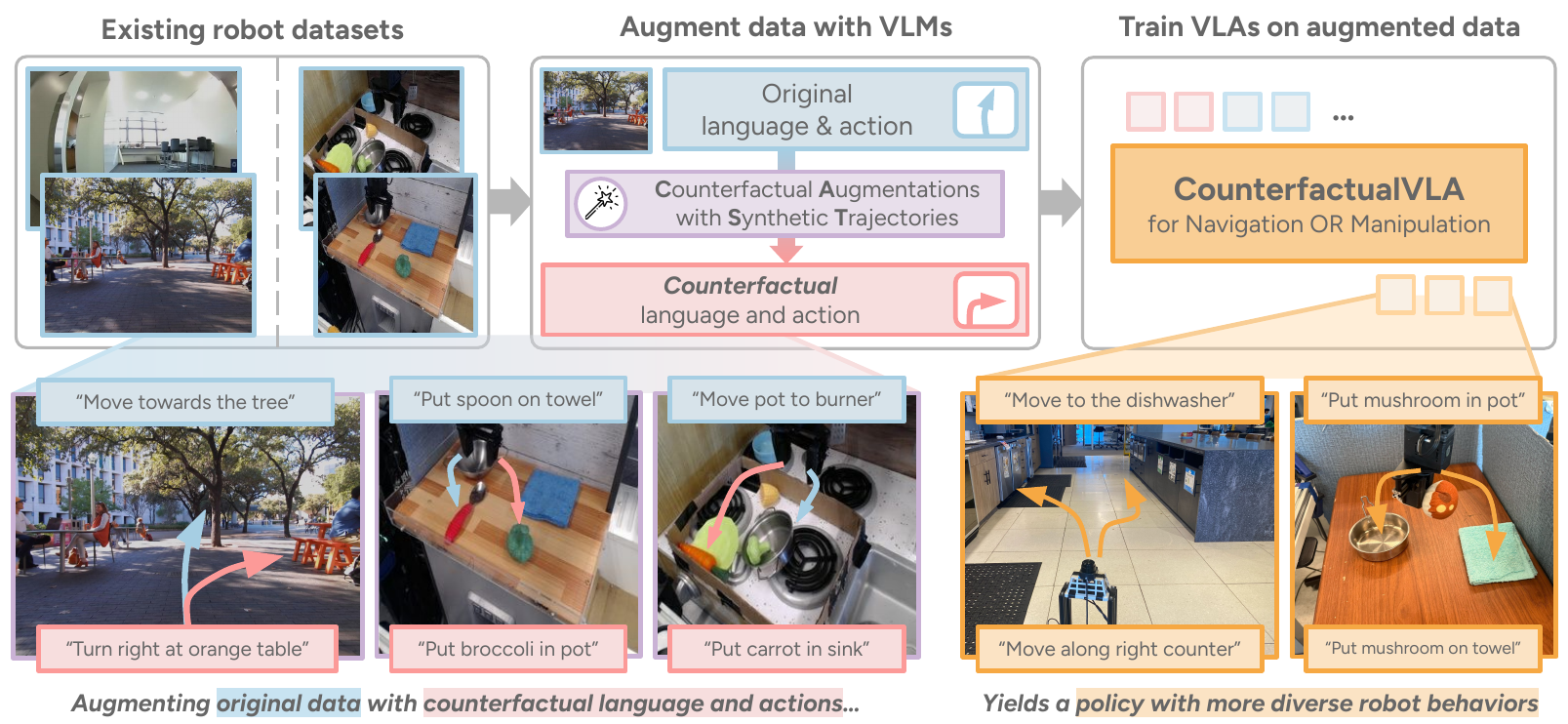}
  \end{center}  
    \vspace{-0.5em}
    {\small Fig. 1:
    \textbf{\MethodName enables better language-following in VLA policies. } \MethodName generates counterfactual actions and language for uncurated robot trajectory datasets using off-the-shelf VLMs. We use this augmented dataset to train CounterfactualVLAs: policies for navigation and manipulation} that can follow complex language instructions in the real world.
    \vspace{-1em}
  \refstepcounter{figure}
  \label{teaser:overview}
}

\makeatother

\maketitle

\setcounter{figure}{1}

\begin{abstract}
Generalist robots should be able to understand and follow user instructions. Despite providing a powerful architecture for mapping open-vocabulary language instructions to robot actions, current vision-language-action (VLA) models struggle to follow fine-grained commands. One cause for this is a lack of semantic diversity and language grounding in existing robot datasets and, specifically, a lack of fine-grained \emph{task} diversity for similar observations. To address this, we present a novel method to \emph{augment} existing robot datasets by leveraging vision-language models to create \emph{counterfactual} labels. By augmenting existing datasets with these labels, we increase the diversity and granularity of language grounding for robot datasets, ultimately improving the language-following capabilities of VLAs.
We evaluate the resulting model's ability to follow language instructions, ranging from simple object-centric commands to complex referential tasks, by conducting vision-language navigation experiments in 3 different indoor and outdoor environments. Our experiments show that counterfactual relabeling (without additional data collection) significantly improves instruction-following in VLA policies, outperforming state-of-the-art methods and doubling the success rate compared to VLAs trained on unaugmented data. We also evaluate our method for manipulation VLAs and find a similar gain in performance on tasks with distractors.
\end{abstract}

\section{Introduction}

A core goal of robotic learning is to enable generalist systems that can perform a diverse set of behaviors in varying scenarios. A major challenge in this effort is training steerable policies -- just as fine-grained instruction-following has been critical in the adoption of image generators~\citep{betker2023dalle3} and language models~\citep{ouyang2022traininglanguagemodelsfollow}, robots should be able to flexibly follow diverse language commands. Vision-language-action models (VLAs) offer one path toward this goal~\citep{brohan2023rt}. Fine-tuned from powerful pre-trained VLMs, VLAs can map open-vocabulary instructions to corresponding robot actions. 
However, their instruction-following abilities are limited by the range of labels in their training corpora. Enabling VLAs to follow more diverse instructions involves scaling up expensive, tedious data curation efforts.
Additionally, when language is vague or not diverse, there is often a stronger correlation between observations and actions (e.g., for an image of a chest of drawers, the most likely task is ``open the drawer''). This leads to \emph{posterior collapse}~\citep{bowman2016generating, he2018lagging}, wherein the policy learns to ignore language, instead inferring the task and actions purely from observations. Some datasets, such as those used for navigation~\citep{shah2022gnm}, lack language labels entirely. As such data tends toward generic navigation behaviors, na\"{\i}vely attaching language labels makes downstream policies especially susceptible to posterior collapse.

In this paper, we address the problem of training steerable language-conditioned robot policies. We propose a method to augment existing robot datasets with synthetic \emph{counterfactual} language labels and behaviors. We first use a VLM to propose alternative instructions a robot could follow at various states in the existing dataset. For example, given a trajectory of a robot going straight down a walkway, as in~\cref{teaser:overview}, our method can generate counterfactual labels such as ``turn right at the orange table'' or ``drive along the tables on the left.''
We then synthesize plausible actions that \emph{would have been} taken in these states to follow these hypothetical instructions, thereby expanding the training language and behavior distribution beyond the original dataset. As generating such actions directly is difficult, we use a VLM to propose atomic commands that correspond to the new instructions (e.g., ``move left'' or ``go straight''). These commands are mapped to actual robot actions by an atomic policy, which generates action chunks that exhibit the desired atomic behavior and branch off from the existing trajectory as counterfactual endings. Finally, the counterfactual language and action data are used to train VLA policies that understand and respond to a broader range of user instructions at test time.

We propose \textbf{C}ounterfactual \textbf{A}ugmentation with \textbf{S}ynthetic \textbf{T}rajectories (\MethodName): a novel data augmentation scheme for improving instruction-following in VLA models. We first instantiate \MethodName for the task of \emph{last-mile} vision-language navigation (VLN), focusing on following nuanced short-horizon prompts (lasting a few seconds or minutes) in unseen environments (e.g., ``Move along the wall on the left'').
We fine-tune an open-source VLM~\citep{beyer2024paligemmaversatile3bvlm} with our augmented dataset, which we call \ModelName, and evaluate performance in \NumEnvs environments on \NumPrompts prompts. We also compare \ModelName to prior methods and find that it outperforms VLAs trained without \MethodName by \DiffCAST\%. We also instantiate CAST for manipulation, focusing on pick-and-place tasks. We train a VLA using the same VLM backbone as for navigation and evaluate on a focused set of tasks with distractors. We find a 27\% performance improvement over a VLA trained with standard data.
We open-source the \MethodName augmentation code, data, and \ModelName training code and checkpoints. We believe that our work is a first step toward truly open-set robotic language following. Code is available here: \href{}{https://github.com/catglossop/CAST}
\section{Related Work}
\label{sec:relatedworks}

\textbf{Task-conditioned policies.} Robot policies often use image-goal conditioning \citep{shah2023vint, sridhar2023nomad, zhu2016targetdrivenvisualnavigationindoor, meng2020scalinglocalcontrollargescale, hirose2019deepvisualmpcpolicylearning}.
However, this requires prior access to the environment or domain-specific image generation models \citep{black2023zeroshotroboticmanipulationpretrained} to produce goals. Language is a more user-friendly way to specify navigation tasks~\citep{gu2022vision, gadre2022cow, shah2022lmnav, huang23vlmaps}. While some methods use only language encoders to obtain language embeddings~\citep{mees2022matterslanguageconditionedrobotic, Gadre2022CLIPOW, hong2021recurrentvisionandlanguagebertnavigation, pashevich2021episodictransformervisionandlanguagenavigation}, VLMs can provide strong language \emph{and} vision priors. Vision-language-action (VLA) policies are fine-tuned on robot data \citep{brohan2023rt, open_x_embodiment_rt_x_2023, kim24openvla, cheng2024navila, chiang2024mobility} to inherit these priors. 
Our goal is to improve VLAs' language-following by training on more diverse synthetic labels. This is \emph{complementary} to architectural and training strategy changes.

\textbf{Zero-shot robot control with VLMs.} An alternative approach is to directly control a robot with a VLM zero-shot~\citep{huang2023voxposer, nasiriany2024pivotiterativevisualprompting, liang2023codepolicieslanguagemodel, sathyamoorthy2024convoicontextawarenavigationusing, gadre2022cow, shah2023lfg, dorbala2023embodied}, thereby obviating the need for language labels during training. However, such methods do not incorporate embodied robot experience during training, relying instead on VLM priors from pre-training tasks such as static image understanding, visual question answering (VQA), object detection, and captioning~\citep{openai2024gpt4technicalreport, geminiteam2024geminifamilyhighlycapable, beyer2024paligemmaversatile3bvlm}. VLMs thus often struggle to translate language prompts into robot actions. \MethodName instead leverages VLM priors by casting counterfactual instruction generation as a captioning task. We query for both counterfactual language and atomic commands, which are translated into action chunks with a policy trained purely on real robot data. As a result, we can effectively ground these synthetic trajectories in realistic behaviors. 

\textbf{Robot data augmentation and reasoning.} \MethodName is closely related to recent advances in improving steerability of language-conditioned robot policies in manipulation and navigation, by using dense hindsight labeling guided by vision foundation models~\citep{cheng2024navila, hirose24lelan, smith2024steerflexibleroboticmanipulation, blank2024scaling} or by using simpler low-level policies in conjunction with foundation models to augment the dataset~\citep{zhang2023bootstrap, ha2023scalingup}. Most related to our work, NaVILA~\citep{cheng2024navila} proposes to train VLAs on synthetic language labels obtained by labeling robot navigation data with VLMs. Our approach differs in that we use a novel \emph{counterfactual} labeling scheme that produces new labels for behaviors that are possible in observed scenes, but were not actually executed. While recent work~\citep{seneviratne2026chopcounterfactualhumanpreference, zhou2026coinscounterfactualinteractivenavigation, peng2025counterfactualvlaselfreflectivevisionlanguageaction} also leverages counterfactual data or reasoning to improve obstacle avoidance or path planning, \MethodName introduces a scalable pipeline for improving \emph{general} language-following in language-conditioned policies.
\begin{figure*}
\vspace{0.5em}
    \centering
    \includegraphics[width=\linewidth]{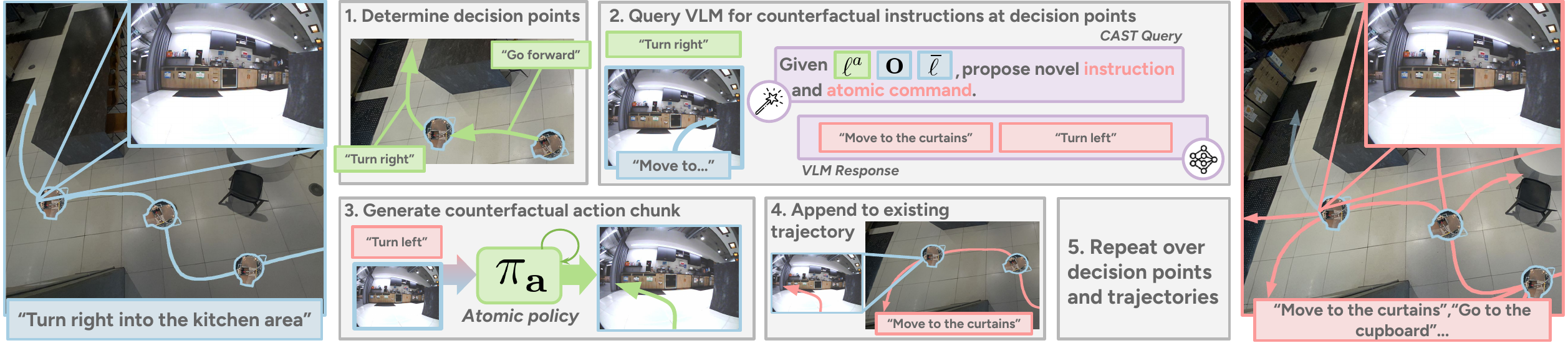}
    \caption{\textbf{An overview of \MethodName.} \MethodName determines decision points in existing trajectories by segmenting them into atomic chunks, forming $\Da$. Then, a VLM is queried with $\bo_{t,i}$ at the decision point and the existing language labels $\lr_{i,j}$ to generate counterfactual instructions $\ell^{cf}_{i}$ and atomic commands $\lacf_{i}$. The corresponding counterfactual action chunks are sampled from $\pia(\bacf_{t:t+H,i} | \lacf_{i}, \bo_{t,i})$ and are appended to the existing trajectories as counterfactual endings.
    }
    \label{fig:method}
    \vspace{-2em}
\end{figure*}

\section{Preliminaries and Problem Statement}
\label{sec:prelim}
Our goal is to train a robot policy to follow complex commands from an unlabeled (or sparsely-labeled) dataset, $\Du = \{ \bo_{t,i}, \ba_{t,i} \}$, where $\ba_t$ denotes the action at time $t$, $\bo_t$ denotes the observation (e.g., an image from the robot's camera), and $i$ denotes $i^\text{th}$ trajectory. To do this, we propose a way to construct a new \emph{densely-labeled} dataset $\Dl = \{ \bo_{t,i}, \ell_{i}, \ba_{t,i} \}$, where $\ell$ denotes a language instruction, e.g., ``move along the white wall.''
However, if these language labels $\ell_{i}$ are not necessary for the model to predict the correct action, then the model will ignore them --- a phenomenon known as \emph{posterior collapse}~\citep{bowman2016generating, he2018lagging}. We aim to attach dense labels such that the policy will listen to them when trained with \emph{imitation learning} (IL), where the model learns to copy the action distributions of experts. 
As we will argue, diverse \emph{counterfactual behaviors}, consisting of language labels $\lcf_{i}$ \emph{and} actions $\bacf_{t,i}$, are an appropriate way to induce this.

We train our policy $\pi_\theta(\ba_{t:t+H} | \ell, \bo_t)$ with parameters $\theta$ to predict \emph{action chunks}, that is short ``trajectory snippets'' over a horizon $H$, which makes it easier to produce temporally-consistent actions and
reduces compounding error~\citep{zhao2023learning, pertsch2025fastefficientactiontokenization}.

\section{Counterfactual Label Augmentation}
\vspace{-0.2em}
\label{sec:method}
In order for policies to be \emph{steerable}, they must pay attention to user-specified language instructions, avoiding posterior collapse. The policy must learn that \emph{different instructions must lead to different actions, even when the observations are otherwise similar}. The language we attach to trajectories must correlate with the action \emph{even when conditioning on the observation}, otherwise the model may ignore the instruction and still achieve high training accuracy. In practice, this is difficult to ensure, as the correlation between the observation and the task is often high -- e.g., the robot might only be asked to navigate to a door when a door is visible.

\MethodName's goal is to attach multiple action-language pairs for each observation $\bo_{t,i} \in \Du$, such that policies must attend to the instruction in order to produce the right action. To do this, we generate potential synthetic language commands $\lcf_{i}$ that are feasible from the observations $\bo_{t,i}$, but are counterfactual to the existing trajectories and labels, $\ell_{i}$, as shown in Step 2 in~\cref{fig:method}. $\Du$ may lack language labels, in which case we synthetically obtain them as described in~\cref{sec:implementation} (but they could be annotated by humans too).

Our key insight is that synthetic counterfactual action chunks
can be obtained from a simpler \emph{atomic} policy, which is much easier to train than a general instruction-following policy and can be reliable over short horizons. This atomic policy $\pia(\ba_{t:t+H,i} | \la, \bo_{t,i})$ follows simple \emph{atomic} language commands $\la$, such as ``turn left'' or ``turn right''. Intuitively, such a policy is less susceptible to posterior collapse, as these atomic commands strongly correlate to the corresponding actions, no matter the observation. Still, the atomic policy must attend to observations to ensure the commands are \textit{reasonable} -- if the command is ``turn left,'' the action should move left while avoiding collisions (see~\cref{sec:implementation}).

To connect the synthetic instruction $\lcf_{i}$ to this atomic policy, we prompt a VLM to generate both the counterfactual instruction $\lcf_{i}$ and a corresponding atomic command $\lacf_{i}$ at a particular state $\bo_{t,i}$. Then, we use the atomic policy $\pia(\ba_{t:t+H,i} | \lacf_{t,i}, \bo_{t,i})$ to generate a suitable action $\bacf_{t,i}$. This constructs an expanded dataset where, for every trajectory, several counterfactual actions are generated that branch off the original data. When training on this expanded dataset, the policy will encounter several instructions that produce distinct actions, but are associated with the same observation. 

\emph{We avoid requiring counterfactual observations by using action chunking, which allows the policy to be perturbed from states covered in the existing data toward states aligned with the counterfactual task.} While we do not provide explicit supervision in these states, we find that these perturbations can effectively expand the language-following capabilities of the policy to novel fine-grained tasks. 

\subsection{How counterfactuals enable language following}
\label{sec:theory}
We motivate our approach by presenting a simple conceptual model for how well labels induce language following. For readability, we drop step and sample subscripts $t$ and $i$, denote the language label as $\ell$, the atomic language label as $\la$, the action as $\ba$, and the observation as $\bo$. $\ell$, $\la$, and $\ba$ could correspond to an original or counterfactual label. 

We expect a model to attend to language input when it adds information about the action $\ba$ that cannot be inferred from the observation $\bo$. This can be captured by the \emph{observation-conditioned} mutual information between the label and the action $I(\ba; \ell | \bo)$. We assume that language-following can be improved by maximizing this quantity, which we argue is a simple and reasonable conceptual model of language following. We will show that by increasing the number of different $\la$ executed at each $\bo$, while ensuring that each atomic command corresponds to a unique $\ell$, we effectively maximize a lower bound on $I(\ba; \ell | \bo)$. This motivates our method as a principled way to improve language following, albeit under this mutual information assumption.

At a given observation $\bo$ in a trajectory, we have several branching action chunks -- one from the original dataset, the others counterfactually generated. Each chunk has an instruction, $\ell$, which describes the actions, $\ba$, taken by the robot at $\bo$. From $\ba$, we can determine an atomic command $\la$, using the heuristics from \cref{sec:implementation}. $\ba$ can also be determined from $\ell$ (this is the mapping learned via imitation learning). We can thus describe this process with the Markov chain $\ell \rightarrow \ba \rightarrow \la$. Applying the data processing inequality yields:
\begin{equation}
    I( \ba ; \ell \vert \bo) \geq I(\la; \ell \vert \bo)
\label{eq:a_MI}
\end{equation}
Expanding the right side, we get
\begin{equation}
\begin{split}
    I(\la ; \ell \vert \bo) &= H(\la \vert \bo) - H(\la \vert \ell, \bo) \\ 
\end{split}
\label{eq:bound}
\end{equation}
\begin{lemma}
\label{eq:lemma_lang}
     Given a language label $\ell$, an atomic command $\la$, and an observation $\bo$, the mutual information $I(\ba ; \ell \vert \bo)$ is lower bounded by $H(\la \vert \bo) - H(\la \vert \ell, \bo)$. 
\end{lemma}
\vspace{2mm}

From Lemma~\ref{eq:lemma_lang}, maximizing $H(\la \vert \bo) - H(\la \vert \ell, \bo)$ optimizes a lower bound on the mutual information between language and action, thereby improving language following.
Thus, an effective way to improve language following is to produce a high-entropy distribution over atomic labels $\la$, while ensuring that the atomic labels are easy to predict given the full language command $\ell$. \MethodName approximates just this: each observation gets multiple counterfactual commands, and thus diverse atomic commands (increasing $H(\la \vert \bo)$), but each counterfactual label has a unique atomic command (decreasing $H(\la \vert \ell, \bo)$). This pushes up on the bound on the mutual information between language and actions, thus improving language following. While the precise relationship between \MethodName and these quantities is heuristic, this formulation provides a degree of conceptual grounding for our approach, motivating relabeling each observation with diverse atomic labels $\lacf$ that are predictable from the corresponding counterfactual labels $\lcf$.

\subsection{Generating the \MethodName dataset}
\label{sec:counterfactuals}

To generate counterfactual language and actions, we require labels associated with the trajectories \emph{actually} executed by the robot as a reference. We construct an atomic decomposition of our target dataset, $\Da = \{ \bo_{t,i}, \la_{t,i}, \ba_{t,i} \}$, as shown in Step 1 of~\cref{fig:method}, and synthetically generate the instructions associated with the trajectories \emph{actually} executed by the robot, $\lr_{i}$. These components are domain-specific and dataset-specific. We provide instantiations of the atomic decomposition and a method to synthetically post-hoc label data with language in~\cref{sec:implementation}.

\textbf{Generating counterfactual instructions.} At a given observation $\bo_{t,i} \in \Du$, we have language that describes the instruction the robot was following ($\lr_{i}$) and the short-horizon atomic robot behaviors in the original trajectory ($\la_{t,i}$).

We use this data as a reference to generate plausible \emph{alternative} language commands $\lcf_{i}$ that the robot \emph{could} have executed, which will correspond to the existing trajectory up to $\bo_{t,i}$ followed by a new ending. We select decision points -- the boundaries of atomic segments in $\Da$ -- where we query a large, API-based VLM for alternative instructions via a \emph{relabeling prompt}. This prompt is constructed using image and atomic label tuples $(\bo_{t_i, i}, \la_{t_i, i})$ at each decision point, as well as the instruction labels $\{\lr_{i}\}$ (step 2 of \cref{fig:method}). It instructs the VLM to describe both the counterfactual behavior that could be executed at the decision point and the \emph{atomic} instruction aligned with this behavior. E.g., at a point where the robot previously turned right toward a hallway, it may be possible for it to instead turn left, corresponding to going toward a door. This yields both an alternative instruction $\lcf_{i}$ and an associated atomic command $\lacf_{i}$ for observation $\bo_{t, i}$. Multiple such \emph{counterfactuals} are generated, potentially for the same observation, approximately raising $H(\la \vert \bo)$. By ensuring each $\lacf_{t,i}$ is associated with a distinct language command, we decrease $H(\la \vert \ell, \bo)$. Thus, we push up on the lower bound on $I(\ba ; \ell \vert \bo)$ from Lemma~\ref{eq:lemma_lang}. The last piece is then to obtain the associated counterfactual actions, $\bacf_{t,i}$.

\textbf{Generating counterfactual actions.} Given counterfactual instructions $\lcf_i$ and their atomic commands $\lacf_i$, we now generate the underlying counterfactual actions. Using $\Da$, we train an atomic policy $\pia( \ba_{t:t+H, i} \vert \la_{t,i}, \bo_{t,i})$. We then sample an action chunk $\bacf_{t,i} \sim \pia(\ba_{t:t+H, i} | \la_{i}, \bo_{t,i})$ for each $(\bo_{t,i}, \lacf_{i})$, yielding a \emph{counterfactual ending} associated with $\lcf_{i}$ that branches from the existing trajectory. Thus, we have the training tuple $(\bo_{t,i}, \lcf_{i}, \bacf_{t,i})$ (though $\bo$ only exists up to $T-h$, where $T$ is the total number of steps and $h$ is the action chunk horizon) that we can add to the labeled training set $\Dl$, where $t$ spans from the beginning of the original trajectory to the counterfactual ending. With this, we have the diverse actions $\bacf$ associated with diverse $\lcf$ for each $\bo$ that push up on the lower bound on $I(\ba; \ell \vert \bo)$ described in Lemma~\ref{eq:lemma_lang}.

\section{Implementation}
\label{sec:implementation}

We instantiate \MethodName for (1) navigation with a ground robot and (2) manipulation with a low-cost robotic arm. This allows us to evaluate the generality of the overall approach. While the core design of \MethodName is the same in both domains, the distinct properties of navigation and manipulation require a few implementation changes. We first describe \MethodName for navigation, then cover the changes for manipulation.

\subsection{Navigation}

We first instantiate \MethodName for vision-language navigation. \MethodName requires a decomposition of atomic behaviors and sparse language labels as reference points for generating counterfactuals. However, large real-world navigation datasets lack text labels. To address this, we create a hindsight relabeling pipeline and a heuristic decomposition of our dataset to determine decision points for generating counterfactuals and training our atomic policy.

\textbf{Atomic labeling and policy.} For each observation $\bo_{t,i} \in \Du$, we first compute an atomic label $\la_{i}$ in the set \{\texttt{turn right}, \texttt{turn left}, \texttt{adjust right}, \texttt{adjust left}, \texttt{go forward}, \texttt{stop}\}, to create the intermediate dataset $\Da = \{\bo_{t,i}, \la_{i}, \ba_{t,i}\}$. We use a simple heuristic to segment trajectories based on when the robot's yaw rapidly changes, then attach appropriate corresponding language labels. If the robot does not turn significantly, the chunk is labeled as ``go forward'' or ``stop'' based on the distance traveled. More detail is provided in Appendix~\ref{app:atomic}. 

This gives us the training set for learning the atomic policy $\pia(\ba_{t:t+H,i} | \la_{t,i}, \bo_{t,i})$ and the decision points for applying our relabeling prompt (step 2 in~\cref{fig:method}). Since the atomic labels $\la_{t,i}$ are computed directly from the actions $\ba_{t,i}$, they correlate strongly with the action, so $\pia(\ba_{t:t+H,i} | \la_{t,i}, \bo_{t,i})$ follows these atomic language labels very well. The policy also inherits good navigation behavior, such as collision avoidance, from the real-world navigation data it is trained on.
The observation is encoded with an EfficientNet-b2~\citep{tan2020efficientnetrethinkingmodelscaling} ConvNet, and the commands $\la$ are embedded with a T5 language encoder~\citep{ni2021sentencet5scalablesentenceencoders}. The image and language embeddings are concatenated and encoded into a context vector, which is used to train an action-chunked diffusion model to predict actions~\citep{sridhar2023nomad}, resulting in the atomic policy $\pia$.

\textbf{Hindsight relabeling pipeline.} We start with a diverse collection of existing visual navigation datasets~\citep{shah2022gnm}. We generate multiple possible language instructions $\lr_{j,i}$ per robot trajectory (indexed by $j$), each describing the entire trajectory. We leverage a VLM to generate these instructions, specifically Gemini 2.5 Pro~\citep{comanici2025gemini25pushingfrontier}. We first prompt the VLM with a subsampled sequence of observations, $\bo_{1:K:T,i}$, and query it to describe the objects, structures, and relative positions of those objects relative to each other and the robot. We query
the VLM to summarize these descriptions and format the result as \emph{new descriptive instructions} that could have served as the prompt when the robot executed the trajectory. We encourage the VLM to leverage a structured format to simplify this initial generation problem. These labels are generated without using the raw robot actions and capture the \emph{intent} of the robot's interaction with the environment. 

We found it useful to further filter this set of labels in a second labeling phase, where we provide the sequence of atomic labels along the trajectory. The VLM is then tasked with filtering $\lr_{j,i}$ to a set of labels that is consistent with the robot's actions, removing instructions that clearly do not align with the trajectory (e.g., ``Move down the hallway on the right'' would be filtered out if the atomic action sequence indicates the robot turns left). This results in a set of labels $\{\lr_{j,i}\}$ for each trajectory, grounded in robot behaviors. More information is provided in Appendix~\ref{app:vlm}.

\textbf{Training a policy.}
We use \MethodName to generate synthetic data for training a navigation VLA. We run our synthetic annotation pipeline on the GNM dataset mixture~\citep{shah2022gnm}, which contains indoor and outdoor robot trajectories from several different embodiments, using Gemini 2.5 Pro~\citep{comanici2025gemini25pushingfrontier} as the VLM. In total, we generate hindsight labels for 65k trajectories with 320k labels, which we filter to 16k trajectories with 65k labels and 86k counterfactual action chunks. We follow ViNT \citep{shah2023vint} and normalize the action spaces across our datasets, transforming them into Cartesian deltas. A key aspect to note is that the counterfactual actions are predicted by the atomic policy with a horizon of 8 steps. As no new observations are generated for these actions, we only train on chunks of the trajectory for which the observation at the start of the chunk is available. These chunks are encoded as 16 tokens. At test time, a separate PD controller converts these deltas into low-level velocities, which we run at 4 Hz.

With these post-processing steps in place, we train a VLA on the \MethodName dataset. We fine-tune the 3B PaliGemma VLM~\citep{beyer2024paligemmaversatile3bvlm} (described in Appendix~\ref{app:model}) for 40k steps on a v4-16 TPU pod with a batch size of 384 and a learning rate of 1e-4. We add 128 tokens to its tokenizer to represent the binned actions.

\subsection{Manipulation}

To implement \MethodName for manipulation, we only modify the counterfactual atomic command proposer and the atomic policy, $\pia$. We use the Bridge dataset~\citep{walke2023bridgedata}, which already has task-level labels (e.g. ``put the corn in the pot''). Then, we use the subtask decompositions open-sourced by Steerable Policies~\citep{chen2026steerablevisionlanguageactionpoliciesembodied}, which serve as atomic commands for manipulation (e.g., ``pick up the corn'' or ``move the corn to the pot''). We use the pipeline from~\cref{sec:counterfactuals} with a modified prompt to generate counterfactual tasks, $\lcf_{i}$, e.g., if the task is ``put the corn in the pot'' and a plate is also present, then a counterfactual task can be ``put the corn on the plate''. Information on the prompts used is provided in Appendix~\ref{app:vlm}.

We also use two models from Steerable Policies~\citep{chen2026steerablevisionlanguageactionpoliciesembodied} to procure counterfactual atomic commands and actions. The first model takes \textit{counterfactual} tasks (``put the corn on the plate'') and observations to predict atomic subtask commands, $\lacf_{i}$ (``move the corn to the plate''). The second model is a pre-trained Steerable Policy which serves as our atomic policy. It predicts actions, $\bacf_{i}$, based on counterfactual atomic commands. Both models are initialized from Prismatic VLMs~\citep{kim24openvla}.  

We run our pipeline on all 60k Bridge trajectories and use Gemini 2.5 Pro~\citep{comanici2025gemini25pushingfrontier} as the VLM to generate 500k counterfactual instructions and actions. Following past works~\citep{chen2026steerablevisionlanguageactionpoliciesembodied, kim24openvla}, the atomic policy produces non-chunked Cartesian delta actions.

We train our manipulation VLA on this augmented dataset, using the same VLM~\citep{beyer2024paligemmaversatile3bvlm} as the VLN policy in the previous section. We train it end-to-end
for 60k steps on $8\times$H200 GPUs with a batch size of 512 and a learning rate of 1e-4. We add 512 tokens to the VLM's tokenizer to represent the action bins. 

We call these models \emph{\ModelName}s, which we experimentally validate in~\cref{sec:main_result}.
\section{Experimental Evaluations}

\label{sec:eval-setup}
Our experiments evaluate the \MethodName dataset and the downstream VLA policy, \ModelName, aiming to answer:

\begin{itemize}
    \setlength{\parskip}{0cm}
    \setlength{\itemsep}{0cm}

    \item[{\bf Q1}] Does \MethodName produce accurate language labels?

    \item[{\bf Q2}]  Does \MethodName enable more effective language-conditioned policies for navigation and manipulation?

    \item[{\bf Q3}]  How does \ModelName trained with \MethodName compare to state-of-the-art methods?

    \item[{\bf Q4}]  Which policy architecture best leverages \MethodName?
\end{itemize}

\textbf{\MethodName dataset evaluation setup.} To evaluate the label quality of \MethodName for \textbf{Q1}, we have 11 human judges evaluate 20 hindsight ($\lr_{j,i}$) and 20 counterfactual labels ($\lcf_{i}$) each, resulting in a total of 440 labels, from the augmented GNM dataset. We evaluate three label aspects: accuracy of the \textit{object(s)/structure(s)}, \textit{motion}, and \textit{grounding} of the motion in the scene. A label is deemed correct if it is accurate with respect to these three aspects, and incorrect otherwise. More detail is provided in Appendix~\ref{app:label_quality}.

\textbf{\ModelName evaluation setup.} For \textbf{Q2}, we evaluate \ModelName on 27 challenging instruction-following navigation tasks across 3 real-world environments, spanning office hallways, a kitchen, and an outdoor public park (see \cref{fig:filmstrip}).
The instructions fall into three categories: \textit{Object Navigation}, where the policy must navigate to within 0.5 m of a target object, \textit{Referential Navigation}, where it must move relative to an object/structure, requiring spatial understanding (e.g. ``Go to the left of...''), and \textit{Continuous Navigation}, where it must perform a continuous behavior relative to objects/structures in the environment, staying within 1 m of them for at least 2 m (e.g. ``Move along the ...''). Except for object reaching, collision fails the trial. See Appendix~\ref{app:nav_tasks} for more details and examples.

These instructions cover a wider range of behaviors than commonly studied in object navigation~\citep{gervet2023objectnav, gadre2022cow, hirose24lelan} and atomic instruction-following literature~\citep{cheng2024navila}, and focus on \emph{steerability} rather than long-horizon navigation. \emph{Steerability is not typically measured in simulation benchmarks because it requires programmatic detectors for sophisticated behaviors, making real-world evaluation more suitable}. We use a Kobuki robot base with a front-facing monocular camera mounted 40 cm above the base. We also mount a LiDAR, which is only used for CoNVOI~\citep{cheng2024navila} evaluations. We perform 5 trials, similar to other works~\citep{cheng2024navila}, totaling over 600 trials and 21 hours of real-world evaluation across all trials.

We also evaluate the robustness of \ModelName with six tasks (two per category) in the hallway setting while varying lighting (morning vs. night) and the presence of dynamic obstacles. This setting has windows, so lighting changes depending on time of day. For the dynamic obstacles, a person walks in front of the robot during each trial.

To evaluate \ModelName for manipulation, we run 10 trials on 6 manipulation tasks on the real-world Bridge WidowX setup~\citep{walke2023bridgedata}. Half are simple ``pick'' tasks (e.g., ``pick up the mushroom'') while the others are more complex (e.g., ``put the mushroom in the pot''). All tasks include \textit{distractor objects}, allowing us to specifically evaluate whether \MethodName improves language following. Without distractors, tasks can be inferred from observations alone, making instruction-conditioning unnecessary. We perform 120 trials across the standard VLA and \ModelName. See Appendix~\ref{app:manip_tasks} for more details and examples. 
\subsection{Evaluating \MethodName label quality}
\label{sec:label_quality_results}

\begin{figure}[t]
    \centering
    \includegraphics[width=\linewidth]{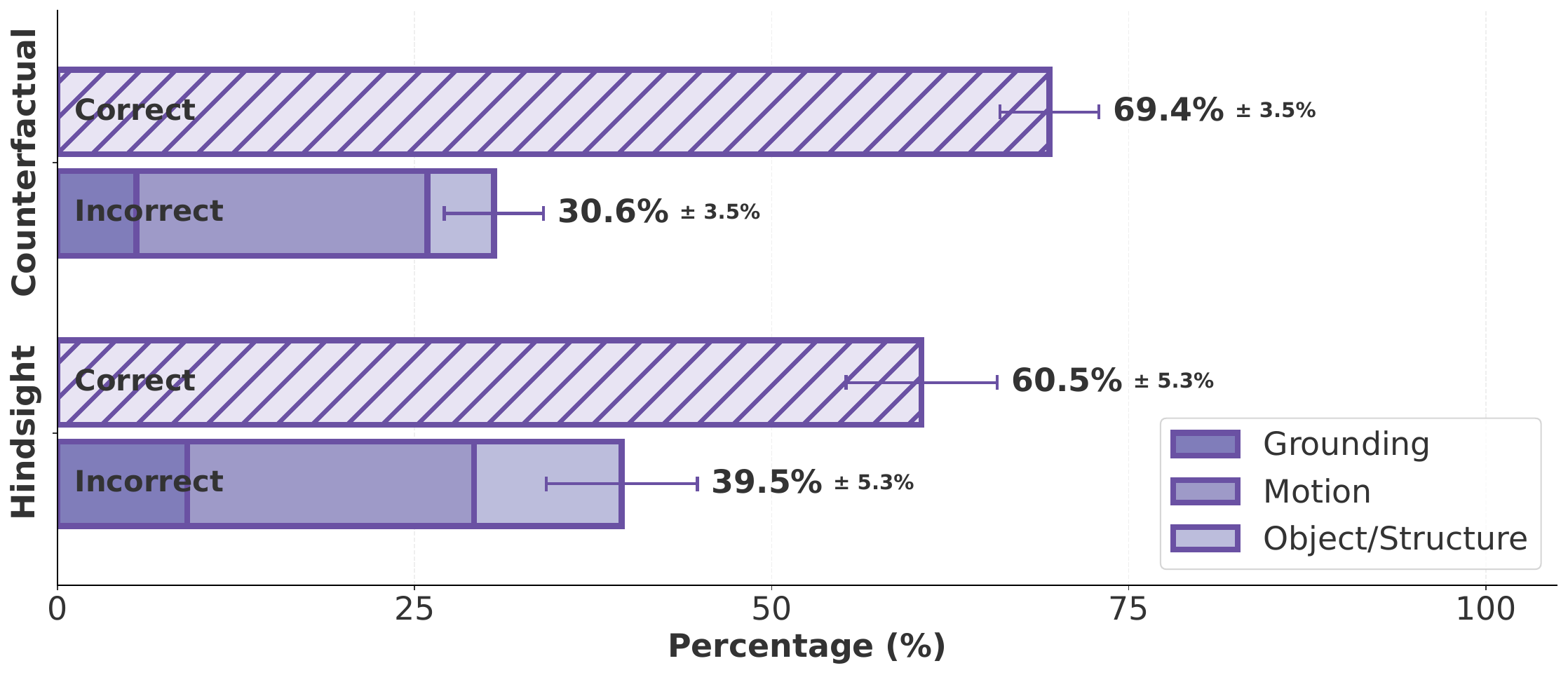}
    \vspace{-1.75em}
    \caption{\textbf{Label Quality.} \MethodName can generate reasonable labels for unlabeled navigation data with an accuracy of 60\% for hindsight (left) and 70\% for counterfactual (right) labels.
    }
    \label{fig:label_results}
    \vspace{-2.0em}
\end{figure}

Toward answering \textbf{Q1}, we find \MethodName generates hindsight and counterfactual labels with approximately 60\% and 70\% accuracy, respectively, as shown in~\cref{fig:label_results}. \MethodName is especially good at recognizing landmark structures or objects in the scene and describing the robot's behavior. The main source of error is grounding of the robot's motion in the scene. For hindsight labels, the VLM often hallucinates the relationship between the motion and the scene (e.g., ``Move along the pathway toward the building'' but the robot leaves the path to go toward the building). For counterfactual trajectories, this manifests as counterfactual endings that would likely result in a collision. 
Still, as shown by our real-world navigation results, \ModelName retains strong collision avoidance behavior when trained with both hindsight and counterfactual labels. Although these labels are noisy under a strict binary evaluation protocol, they still provide sufficient language-action diversity to improve downstream policy learning. 

\begin{figure*}
    \begin{subfigure}[t]{0.5\textwidth}
    \centering    
    \includegraphics[width=\linewidth]{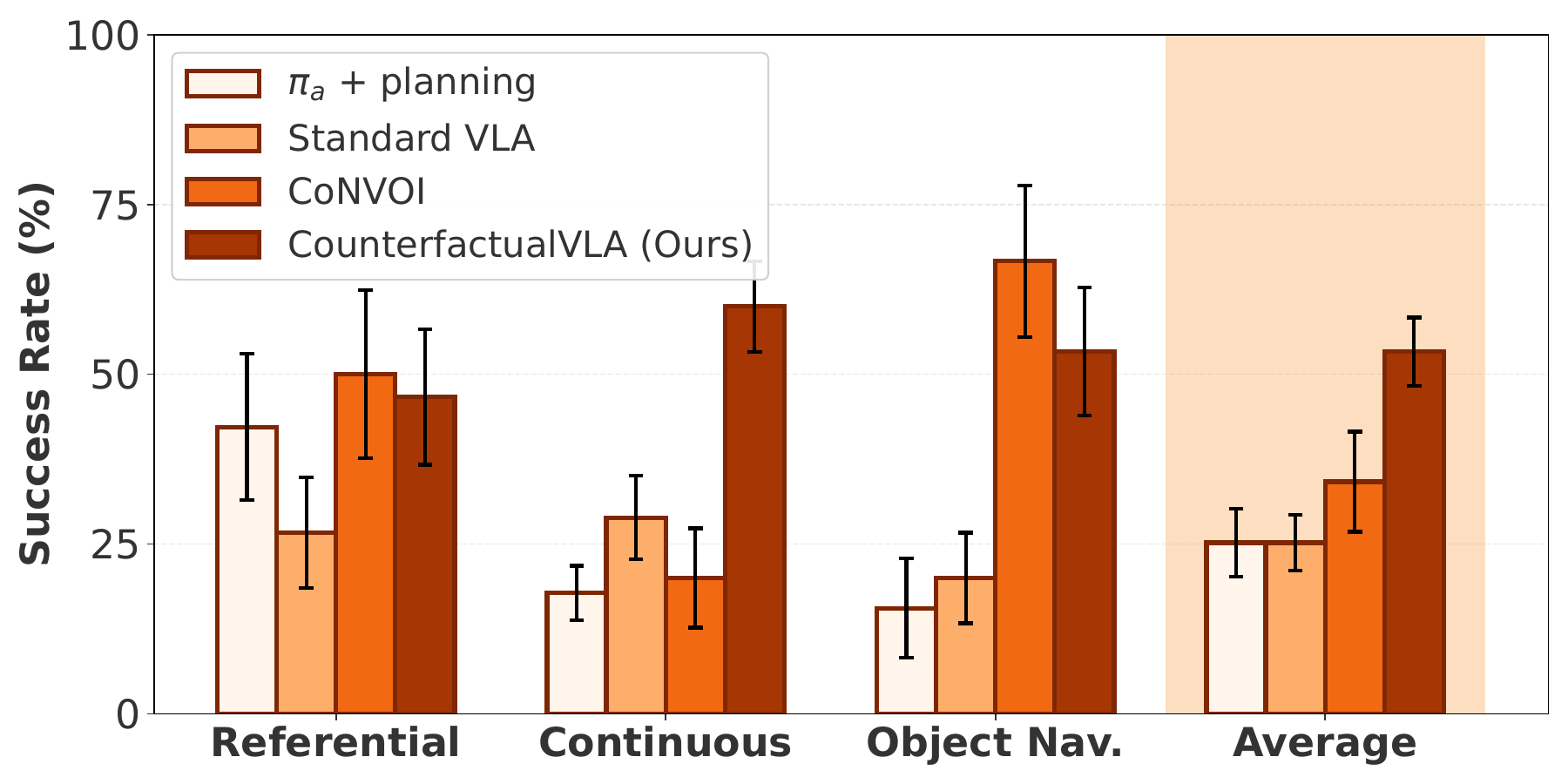}
    \vspace{-1.5em}
    \caption{\textbf{Navigation performance}. \MethodName shows strong performance across a diverse set of tasks, outperforming prior methods.}
    \label{fig:quantitative}
    \end{subfigure}
~
    \begin{subfigure}[t]{0.5\textwidth}
    \centering
    \includegraphics[width=\linewidth]{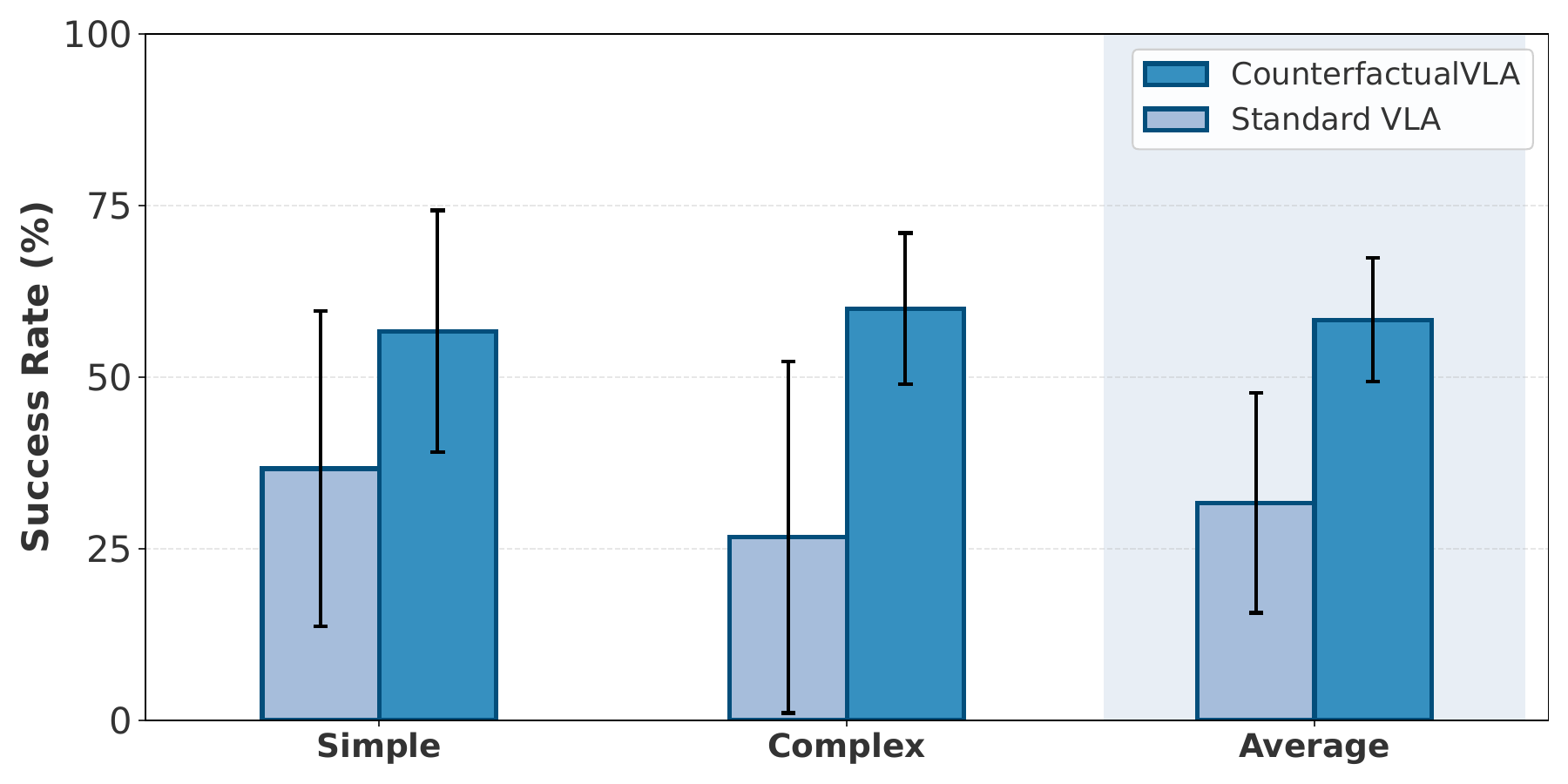}
    \vspace{-1.5em}
    \caption{\textbf{Manipulation performance.} \ModelName doubles the performance of a VLA trained without \MethodName with distractors.}
    \label{fig:manip_results}
    \end{subfigure}
    
\vspace{-0.5em}
\caption{\textbf{Real-world navigation and manipulation evaluations.} \ModelName trained with \MethodName exhibits strong performance across diverse navigation and manipulation tasks. Error bars are $\pm$StdErr.}
\vspace{-0.5em}
\end{figure*}

\begin{figure*}
    \centering
    \includegraphics[width=\textwidth]{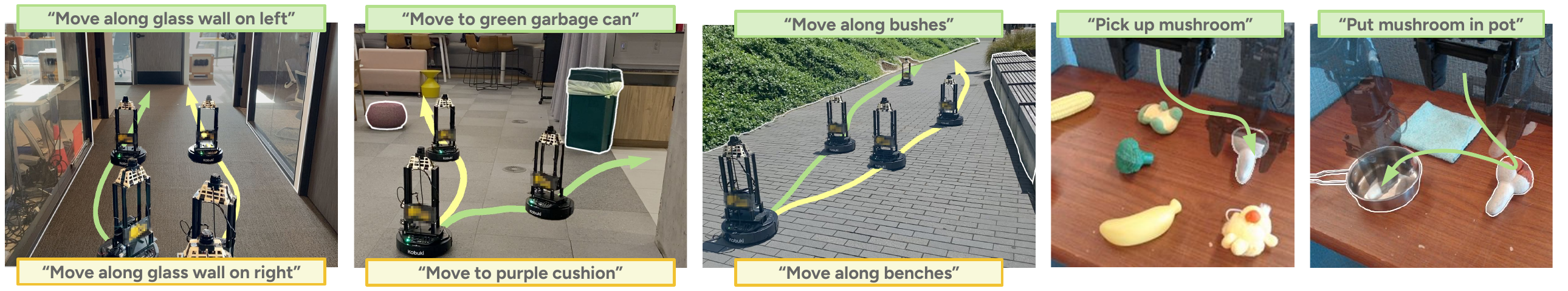}
    \vspace{-1.5em}
    \caption{\textbf{Qualitative examples of \ModelName.} \ModelName can successfully follow complex navigation (left) and manipulation (right) instructions.}
    \label{fig:filmstrip}
    \vspace{-1.5em}
\end{figure*} 

\subsection{Evaluating language instruction following}
\label{sec:main_result}
Our main hypothesis for \textbf{Q2} is that counterfactual labels promote better language following than just the standard labels attached to actions in the original dataset. We test this by comparing \ModelName to a ``standard VLA'', trained with the hindsight task labels alone (without counterfactual labels). These two VLAs are otherwise identical.

As shown in ~\cref{fig:quantitative}, \ModelName achieves a \textbf{27\%} improvement over the standard VLA. Although both policies demonstrate strong navigation capabilities, the standard VLA tends toward more generic navigation behaviors, largely ignoring language. This baseline can succeed where collision avoidance can compensate for a lack of language-following, such as ``move between the rows of chairs'', but struggles in object navigation tasks where collision avoidance can cause failure. The \MethodName dataset uses counterfactual labels to break the bias toward generic behaviors, significantly improving VLAs' ability to follow complex language instructions.

We also find that \MethodName significantly improves instruction following for manipulation (\cref{fig:manip_results}). Without distractors,
both VLAs perform similarly. However, when distractors are introduced, the task is no longer inferable from images alone, necessitating more attention to language. Thus, when the standard VLA sees several plausible behaviors (e.g., placing something in the pot \textit{or} on the cloth), it often oscillates between modes, while \ModelName consistently performs the correct action, achieving 2$\times$ higher success rate.

\subsection{Comparing \ModelName to prior methods}
\label{sec:sota_comparison}

Toward answering \textbf{Q3}, we compare \ModelName against two baselines. First, \textit{$\pia$ + planning}, which follows prior hierarchical frameworks~\citep{cheng2024navila}. We use an off-the-shelf VLM to infer suitable atomic navigation commands for achieving the given language goals, and use the atomic policy $\pia$ to execute them. This investigates whether it is important to distill counterfactual behaviors into our policy or if a VLM can select correct atomic skills at test time. 
Second, \textit{CoNVOI}, which turns instructions into robot plans by annotating image observations with a grid of numbers, prompting a VLM to select a trajectory through this grid, and grounding this plan into the scene geometry with a LiDAR-based occupancy map~\citep{sathyamoorthy2024convoicontextawarenavigationusing}. This requires API calls to a VLM~\citep{openai2024gpt4technicalreport}, assumes that the VLM can express task behaviors with a trajectory of this form, and requires LiDAR for grounding (whereas our VLAs only require images). CoNVOI can accept open-set language, unlike other similar works~\citep{cheng2024navila}, making it the most appropriate baseline for our method. We could not evaluate CoNVOI outside due to infrastructure challenges. More detail on these baselines is provided in Appendix~\ref{app:baselines}.

\ModelName outperforms prior methods by \textbf{19\%} on aggregate (\cref{fig:quantitative}). The best baseline, CoNVOI, achieved similar performance in object and referential navigation. However, it struggled with continuous navigation, where spatial understanding is crucial for avoiding collision while staying close to structures or objects mentioned in the prompt. Since CoNVOI relies on disjoint sources for semantic (VLM) and geometric (LiDAR) information, it struggles to bridge the two, especially in continuous tasks. As \ModelName is trained with embodied robot data, it can better follow nuanced continuous navigation prompts.

The $\pia$ + planning baseline performs poorly on all tasks. The test-time decomposition may appear similar to \MethodName, but $\pia$ + planning cannot know the impact of a particular command until it has been rolled out. Errors in the action chunks themselves are compounded with failures of the VLM to interpret the scene geometry and reason about the next atomic command, causing poor grounding of the robot in the scene and poor adherence to language instructions. By incorporating information across the trajectory, \MethodName achieves tighter alignment between language annotations and grounded robot experience, resulting in significantly stronger language-following capabilities in downstream policies. 

\ModelName is robust to lighting changes but is susceptible to dynamic obstacles, as shown in~\cref{tab:robustness}. While lighting does not seem to impact performance, dynamic obstacles often cause the policy to fail to return to the task after avoiding the obstacle. This may be due to slow inference speed (4 Hz) or the lack of counterfactual observations.

\begin{table}[t]
\centering
\label{tab:conditions}
\begin{tabular}{lccc}
\toprule
\textbf{Method} & \textbf{Time of day} & \textbf{Dynamic obs.} & \textbf{Success (\%)} \\
\midrule
\multirow{4}{*}{\textbf{CounterfactualVLA}} & Morning & No & 63 {\tiny $\pm$ 9}\\
& Morning & Yes & 37 {\tiny $\pm$ 9}\\
& Night & No & 63 {\tiny $\pm$ 9}\\
& Night & Yes & 43 {\tiny $\pm$ 9}\\
\bottomrule
\end{tabular}

\vspace{-0.5em}
\caption{\ModelName is robust to ambient lighting changes but remains susceptible to dynamic obstacles.}
\label{tab:robustness}
\vspace{-3em}
\end{table}

\subsection{Ablating policy architecture for \MethodName}
\label{sec:model_discussion}

\begin{figure}
    \centering
    \includegraphics[width=\linewidth]{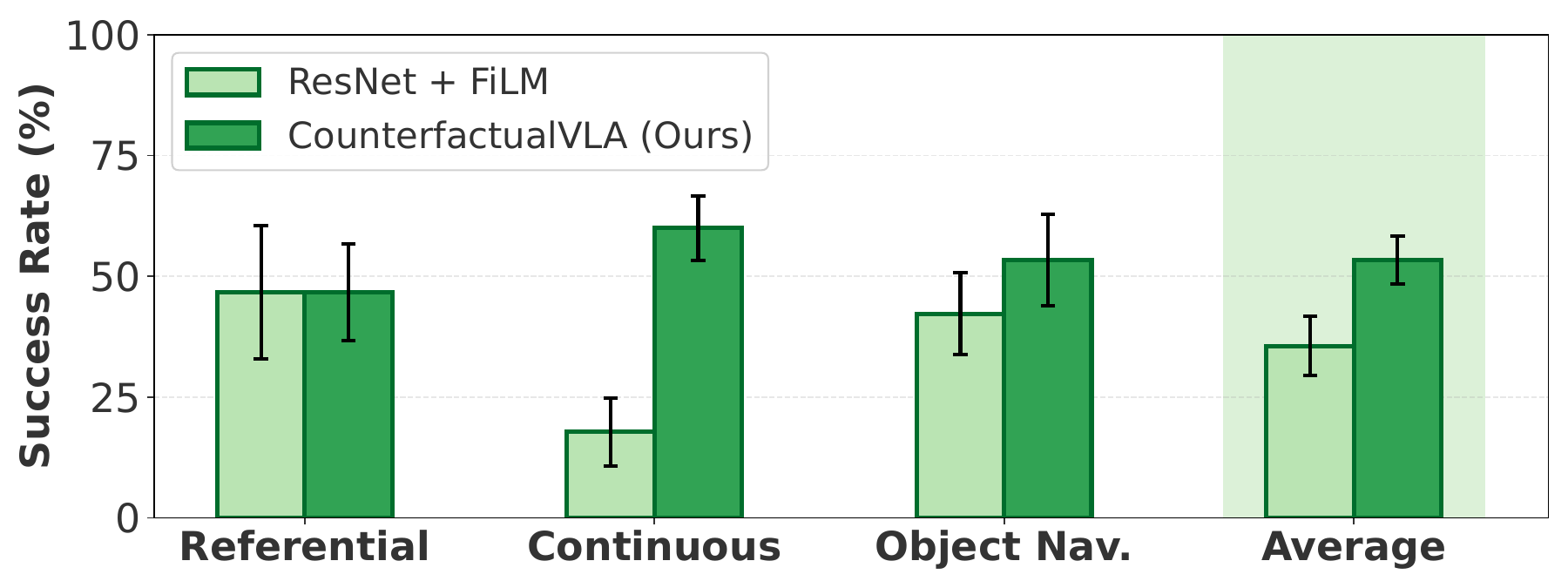}
    \vspace{-2em}
    \caption{\textbf{Model architecture ablation.} VLA backbones offer semantic priors, enabling better complex language following.}
    \label{fig:architecture_ablation_results}
    \vspace{-2em}
\end{figure}

For \textbf{Q4}, we train a policy using ResNet + FiLM (instead of PaliGemma) with \MethodName to analyze the impact of architecture on instruction-following. This architecture is representative of policies used in many prior works~\citep{octo_2023, hirose24lelan}. Following LeLaN~\citep{hirose24lelan}, the language encoder is a frozen CLIP model. \ModelName (using a VLM backbone) exhibits better performance in all tasks (\cref{fig:architecture_ablation_results}). The ResNet + FiLM policy struggled with complex instructions, but was steerable in tasks that could be reduced to object navigation or atomic commands. As \MethodName generates diverse language, a high-capacity model is needed to make full use of it.

\section{Discussion}
\label{sec:discussion}

We present \MethodName, a novel synthetic labeling method that uses VLMs to generate diverse counterfactual labels and robot actions. Even without additional human data, augmenting existing robot datasets with \MethodName enables VLAs that attend more closely to instructions. By assigning \emph{multiple} language-action pairs to each observation, \MethodName encourages policies to learn the link between language and actions. Our method roughly doubles the success rate compared to equivalent standard VLAs trained without counterfactual labels in both navigation and manipulation.

\textbf{Limitations:} 
CAST annotations remain imperfect, though VLMs will likely continue to improve their physical grounding, making our method more effective. 
Counterfactual \textit{observations} would also likely aid our method, but generating robot images introduces new challenges, like ensuring visual consistency with real data or addressing compounding video generation error.
\MethodName shows promise in using \emph{existing} robot datasets to train steerable robot policies. 

\hypersetup{linkcolor=red, urlcolor=blue}
\bibliographystyle{IEEEtran}
\bibliography{references}

\appendix
\section{Appendix}
\label{app:appendix}
\renewcommand{\thesubsection}{\Alph{subsection}}

\subsection{Atomic Discretization of Navigation Dataset}
\label{app:atomic}

We use Cartesian waypoint representations for our robotic navigation task, and since these labels are purely functions of the robot's Cartesian-space trajectory, they can be computed with simple automated heuristics that discretize the trajectory into segments. Each segment is determined by chopping the trajectory at points where the change in yaw exceeds a manually selected threshold, indicating a turn has occurred and labeling it appropriately. If a significant change in yaw does not occur over a set number of steps, in this case 10, then the chunk is labeled as \textit{Go forward} or \textit{Stop} based on the distance traveled. This yields the atomic labels $\la_{t,i}$, along with lists of time steps $\{\ta_{i,j}\}$ indicating the start of each atomic segment (which we will use later in our method). These labels are generated independently of $\bo_{t,i}$, meaning they are strongly correlated with $\ba_{t,i}$.

The Bridge manipulation dataset~\citep{walke2023bridgedata} is already decomposed into subtasks which we use as the atomic discretization of the dataset. These labels were originally produced in the Embodied Chain-of-Thought work~\citep{Zawalski24-ecot}. 

\subsection{Description of Criteria for Label Quality}
\label{app:label_quality}

We use three criteria to determine the correctness of the labels generated by \MethodName:
\begin{itemize}
    \item \textit{Object(s)/structure(s):} Are the object(s)/structure(s) referenced in the label in the scene?
    \item  \textit{Motion:} Is the motion of the robot correctly described? (i.e. If the robot moved right to go toward an object, the label says ``Move right to ...'' rather than ``Move left to ...''). 
    \item \textit{Grounding:} If the motion and objects/structures described correctly, is their interaction described correctly (is the motion properly grounded?) (i.e. if the robot turns left away from an object, the label should be ``Turn left away from <object>'' rather than ``Turn left toward <object>''). Note that for the counterfactual labels, this also includes if the trajectory generated would have led to collision or not. 
\end{itemize}

We ask 11 human judges to evaluate 20 hindsight and 20 counterfactual labels, resulting in a total of 440 labels, deeming a label correct if it is correct with respect to all these criteria and otherwise deem it incorrect. 

\subsection{Description of Baselines}
\label{app:baselines}

\vspace{1mm}
\MyPara{CoNVOI~\cite{sathyamoorthy2024convoicontextawarenavigationusing}:} CoNVOI uses a VLM to process image observations with visual markers denoting future navigation plans. It uses a LiDAR point cloud to generate an occupancy grid, and plans a path through the environment using a combination of VLM guidance and classical planning. While our method does not use privileged observations from a LiDAR, we treat the comparison with CoNVOI as a state-of-the-art method that leverages off-the-shelf VLMs as well as classical planning.

\vspace{1mm}
\MyPara{$\pia$ + planning:} This baseline uses an off-the-shelf VLM to orchestrate a low-level learned policy in a hierarchical control fashion~\cite{cheng2024navila,zhang2023bootstrap,saycan2022arxiv}. This baseline is most similar to NaVILA~\cite{cheng2024navila}, which fine-tunes a VLM to output low-level language commands that are executed by a learned policy. Our version uses a powerful base VLM \cite{geminiteam2024geminifamilyhighlycapable} as the high-level policy and the same atomic policy used for \MethodName as the low-level policy.

The following is the prompt used with the VLM
\definecolor{mytextcolor}{rgb}{0.63,0.08,0.18} 
\definecolor{mybackgroundcolor}{rgb}{0.95, 0.95, 0.95} 
\begin{lstlisting}[
    basicstyle=\ttfamily\footnotesize\color{mytextcolor},
    backgroundcolor=\color{mybackgroundcolor},
    breaklines=true,
    caption=VLM prompt for baseline.,
    label=lst:pia_planning,
]
"vlm_prompt = f'A robot is moving through an environment and has the task '{prompt}'. Given the current observation, which action in the list {PRIMITIVES} should the robot take next? Return your response as the single action in the list of primitives with no additional information.'",
\end{lstlisting}

\subsection{Description of navigation tasks}
\label{app:nav_tasks}

\begin{figure*}[t!]
    \centering
    \begin{subfigure}[t]{\textwidth}
        \centering
        \includegraphics[height=2.5in]{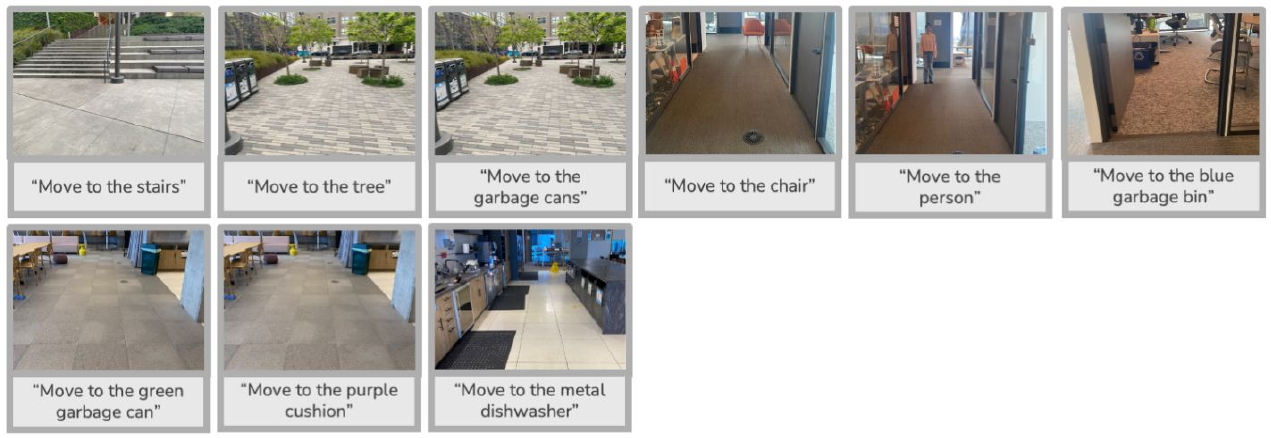}
        \caption{\textbf{Object navigation tasks.} The robot must move toward and reach within 50 cm of a specified object. A collision is okay if it is with the target object.}
    \end{subfigure}%
    \\ 
    \begin{subfigure}[t]{\textwidth}
        \centering
        \includegraphics[height=2.5in]{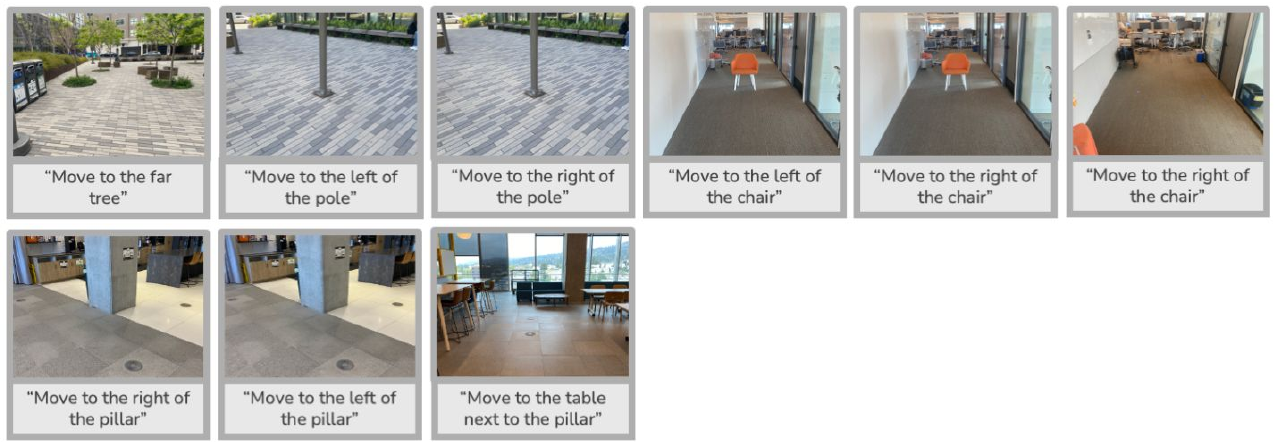}
        \caption{\textbf{Referential navigation tasks.} The robot must navigate relative to an object or structure in the environment, following instructions with directions such as ``move left/right''.}
    \end{subfigure}
    \\
    \begin{subfigure}[t]{\textwidth}
        \centering
        \includegraphics[height=2.5in]{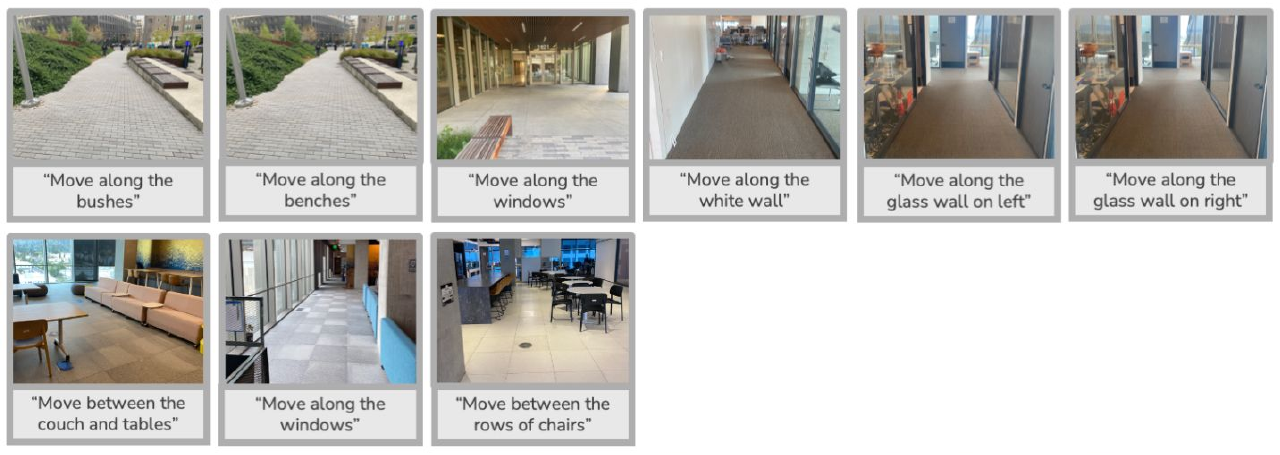}
        \caption{\textbf{Continuous navigation tasks.} The robot must move along a specified structure(s) in the environment and must continue the behavior over a horizon of 2 m.}
    \end{subfigure}
    \caption{\textbf{Navigation tasks.} \ModelName is evaluated in 3 different indoor and outdoor environments across three task types (object navigation, referential navigation, and continuous navigation). In total, \ModelName is evaluated on 27 different navigation tasks.}
    \label{fig:nav_tasks}
\end{figure*}

Refer to~\cref{fig:nav_tasks} for visualizations of the tasks. 

\vspace*{0.5mm}
\noindent\textbf{Object navigation:} The policy is instructed to go to a specific object in the environment. In each of the environments, the objects are placed on opposite sides of the robot's starting position in front of the robot. The rollout is marked as successful if it reaches within 50 cm of the target object. 

\vspace*{0.5mm}
\noindent\textbf{Referential navigation:} The policy is instructed to interact with an object or structure in the environment in a referential way that requires spatial understanding. For example, ``move to the right of the chair.'' or ``move to the table next to the pillar''. The rollout is marked as successful if the policy moves in the correct way without collision or to the correct object.

\vspace*{0.5mm}
\noindent\textbf{Continuous navigation:} The policy is instructed to interact with an object or structure in the environment that requires a continuous behavior. For example, ``move along the wall". The trial is considered successful if the policy moves the robot to within 1 m of the reference structure, for behaviors such as ``move along ...'', and/or continues the behavior for 2 m or more.  

\subsection{Description of manipulation tasks}
\label{app:manip_tasks}

\begin{figure*}[t!]
    \centering
    \begin{subfigure}[t]{0.48\textwidth}
        \centering
        \includegraphics[height=2in]{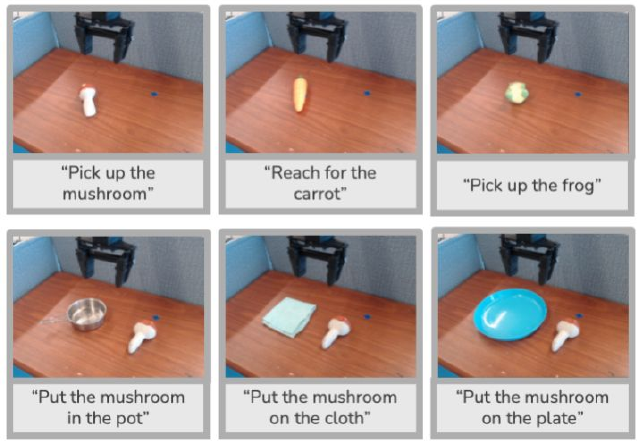}
        \caption{\textbf{Simple manipulation tasks.} The robot must pick up and hold or reach for a specified object.}
    \end{subfigure}%
    ~
    \begin{subfigure}[t]{0.48\textwidth}
        \centering
        \includegraphics[height=2in]{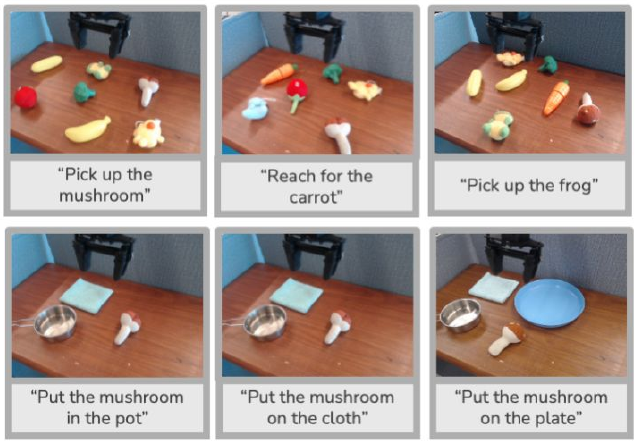}
        \caption{\textbf{Complex manipulation tasks.} The robot must pick up and place a specified object on or in a specified vessel.}
    \end{subfigure}
    \caption{\textbf{Manipulation tasks.} \ModelName is evaluated on manipulation tasks with clutter. We show the same tasks without clutter/distractors on the left side and what the task looks like with distractors on the right. In total, we evaluate \ModelName on 6 different manipulation tasks.}
    \label{fig:manip_tasks}
\end{figure*}

Refer to~\cref{fig:manip_tasks} for visualizations of the tasks.

\noindent\textbf{Simple manipulation:} The policy is instructed to pick up a specific object. The policy must pick up and hold the object for the trial to be a success. 

\noindent\textbf{Complex manipulation:} The policy is instructed to pick up a specific object and put it on or in a specified vessel. For the trial to be successful, the episode must end with the object on or in the vessel.

\subsection{Model Architecture and Training}
\label{app:model}

\begin{figure*}
    \centering
    \includegraphics[width=\textwidth]{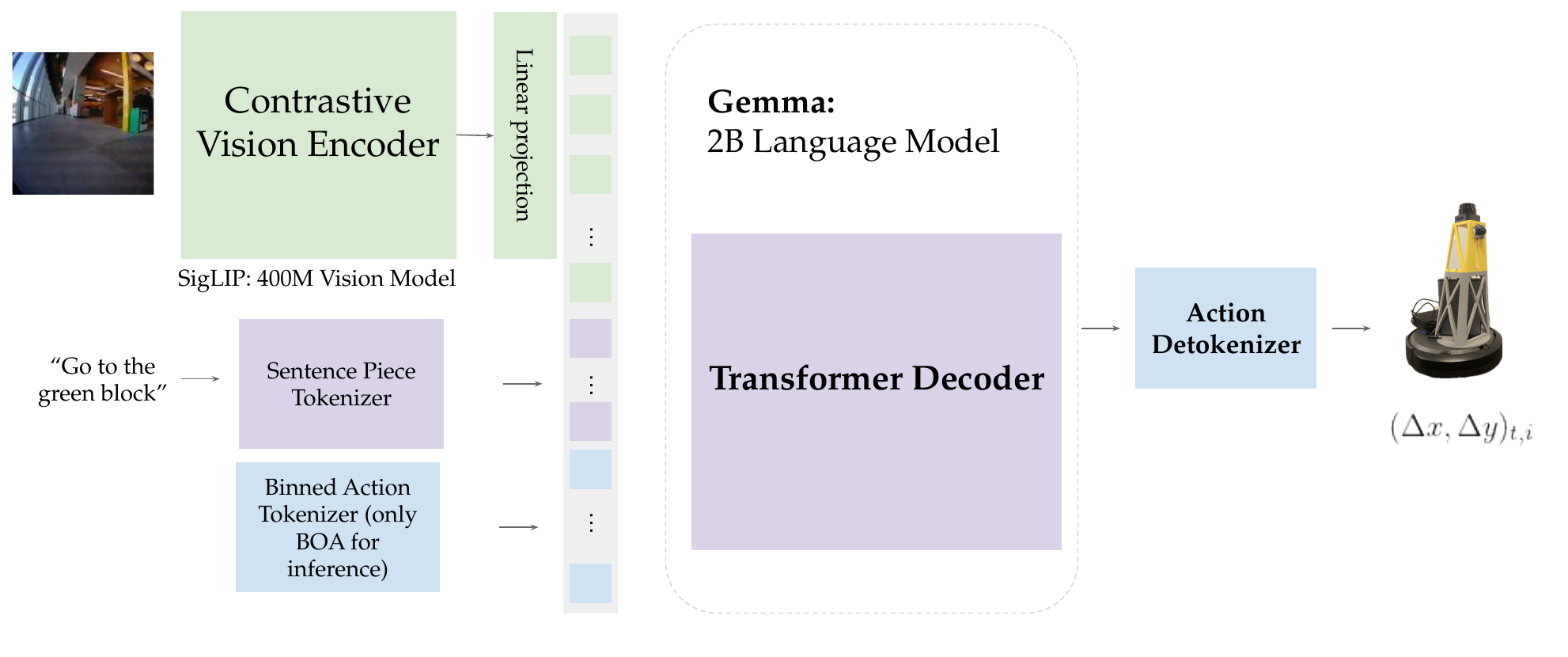}
    \caption{Architecture diagram of the \ModelName, which is based on a PaliGemma VLM backbone. We introduce additional action tokens to ground the VLM in robot actions and fine-tune it with \MethodName data to follow language instructions in the real world.}
    \label{fig:architecture}
\end{figure*}

We provide a more detailed visualization of the \ModelName base VLM architecture in~\cref{fig:architecture}, PaliGemma~\citep{beyer2024paligemmaversatile3bvlm}. PaliGemma is a 3 billion parameter model consisting of a 2B language model (Gemma~\citep{gemmateam2024gemmaopenmodelsbased}) and a 400M parameter SigLIP~\citep{zhai2023sigmoidlosslanguageimage} vision-language model.

Figure~\ref{fig:architecture} shows a schematic of our 3B parameter \ModelName model, which uses a PaliGemma VLM backbone. Our trained checkpoints are available \href{https://huggingface.co/catglossop/CounterfactualVLA}{here}.

\subsection{Detailed Experimental Results}
\label{app:results}
We conducted 645 real-world navigation trials and 120 real-world manipulation trials to evaluate the performance of \ModelName and baselines. The summarized results are available in~\cref{fig:quantitative} and~\cref{fig:manip_results}. We provide the detailed experimental results for navigation (\cref{tab:nav_results}) and manipulation (\cref{tab:manip_results}). In the manipulation results, we additionally provide the performance of the policies on tasks without distractors, corresponding to an additional 120 trials, further demonstrating the performance delta that \MethodName provides for policies in cluttered environments. {Additional visualizations of the policy are available on \href{https://cast-vla.github.io/}{our website}. Note that 6 tasks are missing for CoNVOI's evaluation due to infrastructure challenges. 

\begin{table*}
    \centering
    \tiny
    \begin{tabular}{lcccccc}
         \toprule
         \textbf{Prompt} & \textbf{Environment} & \textbf{CounterfactualVLA} & \textbf{Standard VLA} & \textbf{$\pia$ + planning} & \textbf{CoNVOI} & \textbf{ResNet + FiLM} \\
         \midrule \vspace{2pt}
         \textbf{Object Navigation} & & & & & & \\
         "Move to the orange chair" & Office hallway & 3/5 & 0/5 & 0/5 & 1/5 & 2/5 \\
         "Move to the person" & Office hallway & 4/5 & 1/5 & 1/5 & 4/5 & 1/5 \\
         "Move to the blue garbage bin" & Office hallway & 4/5 & 2/5 & 1/5 & 3/5 & 3/5\\
         "Move to the stairs" & Outdoors & 1/5 & 0/5 & 2/5 & - & 3/5 \\
         "Move to the tree" & Outdoors & 0/5 & 1/5 & 3/5 & - & 3/5 \\
         "Move to the garbage cans" & Outdoors & 2/5 & 1/5 & 0/5 & - & 3/5 \\
         "Move to the green garbage can" & Common area/Kitchen & 4/5 & 0/5 & 0/5 & 4/5 & 1/5 \\
         "Move to the metal dishwasher" & Common area/Kitchen & 3/5 & 1/5 & 0/5 & 5/5 & 0/5 \\
         "Move to the purple cushion" & Common area/Kitchen & 3/5 & 1/5 & 0/5 & 3/5 & 2/5 \\
         \midrule  \vspace{2pt}
         \textbf{Continuous Navigation} & & & & & & \\
         "Move along the glass wall on the left" & Office Hallway & 4/5 & 0/5 & 0/5 & 0/5 & 3/5\\
         "Move along the glass wall on the right" & Office Hallway & 2/5 & 0/5 & 1/5 & 0/5 & 2/5\\
         "Move along the white wall" & Office Hallway & 4/5 & 0/5 & 1/5 & 1/5 & 0/5 \\
         "Move along the benches" & Outdoors & 4/5 & 0/5 & 1/5 & - & 0/5 \\
         "Move along the bushes" & Outdoors & 2/5 & 0/5 & 2/5 & - & 1/5 \\
         "Move along the windows" & Outdoors & 2/5 & 0/5 & 1/5 & - & 1/5\\
         "Move between the pink couch and the tables" & Common area/Kitchen & 2/5 & 0/5 & 1/5 & 2/5 & 1/5\\
         "Move between the rows of chairs" & Common area/Kitchen & 3/5 & 1/5 & 1/5 & 2/5 & 0/5 \\
          "Move along the windows" & Common area/Kitchen & 4/5 & 1/5 & 0/5 & 1/5 & 0/5 \\
         \midrule  \vspace{2pt}
         \textbf{Referential Instructions} & & & & & & \\
         "Move to the left of the chair" & Office Hallway & 3/5 & 1/5 & 1/5 & 0/5 & 2/5\\
         "Move to the right of the chair" & Office Hallway & 2/5 & 0/5 & 1/5 & 0/5 & 1/5\\
         "Move to the door on the right" & Office Hallway & 2/5 & 0/5 & 1/5 & 0/5 & 3/5 \\
         "Move to the left of the pole" & Outdoors & 4/5 & 0/5 & 2/5 & 0/5 & 3/5\\
         "Move to the right of the pole" & Outdoors & 2/5 & 1/5 & 3/5 & 1/5 & 2/5 \\
         "Move to the far tree" & Outdoors & 0/5 & 1/5 & 0/5 & 1/5 & 1/5 \\
         "Move to the left of the pillar" & Common area/Kitchen & 1/5 & 5/5 & 1/5 & 4/5 & 2/5 \\
         "Move to the right of the pillar" & Common area/Kitchen & 1/5 & 5/5 & 1/5 & 4/5 & 1/5\\
         "Move to the table next to the pillar" & Common area/Kitchen & 2/5 & 1/5 & 1/5 & 4/5 & 3/5\\
         \bottomrule
    \end{tabular}
    \caption{Detailed navigation results}
    \label{tab:nav_results}
    \vspace*{-2mm}
\end{table*}

\begin{table*}
    \centering
    \tiny
    \begin{tabular}{lcccccc}
         \toprule
         \textbf{Prompt} & \multicolumn{2}{c}{\textbf{CounterfactualVLA}} & \multicolumn{2}{c}{\textbf{Standard VLA}} \\
         \midrule \vspace{2pt}
         & \textbf{w/o distractors} & \textbf{w/ distractors} & \textbf{w/o distractors} & \textbf{w/ distractors} \\
         \hline
         ``Pick up the mushroom'' & 8/10 & 5/10 & 8/10 & 3/10\\
         ``Reach for the carrot'' & 10/10 & 9/10 & 8/10 & 8/10 \\
         ``Pick up the frog''  & 6/10 & 4/10 & 3/10 & 0/10 \\
         ``Put the mushroom in the pot''  & 3/10 & 8/10 & 8/10 & 0/10\\
         ``Put the mushroom on the cloth'' & 3/10 & 2/10 & 2/10 & 0/10 \\
         ``Put the mushroom on the plate'' & 5/10 & 4/10 & 3/10 & 8/10 \\
         \hline \\ 
         Total & 35/60 & 35/60 & 32/60 & 19/60 \\
         \bottomrule
    \end{tabular}
    \caption{Detailed manipulation results}
    \label{tab:manip_results}
    \vspace*{-2mm}
\end{table*}

\subsection{VLM Relabeling}
\label{app:vlm}

We use state-of-the-art VLMs with commercially available APIs in various stages of our pipeline. For all the counterfactual relabeling and filtering, we use Gemini 2.5 Pro~\citep{comanici2025gemini25pushingfrontier}. Since we do not make any assumptions on the specific capabilities of these models, our proposed method is compatible with any similar VLM and will benefit from any future improvements. We provide the exact prompts used by our method during various stages of the relabeling and filtering process.

The exact prompts we use for labeling navigation data are provided in our \href{https://github.com/catglossop/CAST/tree/main/cast/data/prompts}{repo}.

To apply \MethodName to the Bridge manipulation dataset~\citep{walke2023bridgedata}, we do not need to perform the initial generation of language labels as the data is already labeled with task instructions. We modify the counterfactual generation step to be applied to this data with the following prompt: 

\begin{lstlisting}[
    basicstyle=\ttfamily\footnotesize\color{mytextcolor}, % Text color
    backgroundcolor=\color{mybackgroundcolor}, % Background color
    breaklines=true,
    caption= Counterfactual generation prompt for manipulation data.,
    label=lst:init
]
"A robot arm is performing a task described by any of the original instructions {filtered_language}. So far, it has executed the subtasks {subtasks}. Provided is the current observation of the robot and the scene. Propose a list of new instructions, different from the original instructions that the robot could have done instead given the provided image. This may include interacting with a different object, placing an object in a different way, moving relative to a different object and so on. Each instruction should be concise (1 sentence, 5-15 words) and should be similar in style to the original instructions.",
\end{lstlisting}

\end{document}